\documentclass[10pt,twocolumn,letterpaper]{article}

\usepackage{iccv}
\usepackage{times}
\usepackage{epsfig}
\usepackage{graphicx}
\usepackage{amsmath}
\usepackage{amssymb}
\usepackage{bbding}
\usepackage{algorithm}
\usepackage{amsthm}
\usepackage{multirow}
\usepackage{algpseudocode}
\usepackage{graphics}
\usepackage{epsfig}
\usepackage{xcolor}

\graphicspath{{figs/}}

\usepackage[pagebackref=true,breaklinks=true,letterpaper=true,colorlinks,bookmarks=false,hyperfootnotes=true]{hyperref}

 \iccvfinalcopy 

\def\iccvPaperID{1272} 

\ificcvfinal\fi
\begin{document}
	
	\title{Data-Free Learning of Student Networks}
	
	\author{
	Hanting Chen$^{1}$\thanks{This work was done while visiting Huawei Noah's Ark Lab}  , Yunhe Wang$^2$, Chang Xu$^3$\thanks{corresponding author}, Zhaohui Yang$^{1*}$, Chuanjian Liu$^2$, Boxin Shi$^{4,5}$, \\Chunjing Xu$^2$, Chao Xu$^1$, Qi Tian$^2$\\
	\normalsize$^1$ Key Lab of Machine Perception (MOE), CMIC, School of EECS, Peking University, China\\
    \normalsize$^2$ Huawei Noah's Ark Lab, \normalsize$^3$ School of Computer Science, Faculty of Engineering, The University of Sydney, Australia\\
    \normalsize$^4$ National Engineering Laboratory for Video Technology, Peking University, \normalsize$^5$ Peng Cheng Laboratory\\
	\small\texttt{\{chenhanting, zhaohuiyang, shiboxin\}@pku.edu.cn, c.xu@sydney.edu.au} \\
	 \small\texttt{\{yunhe.wang, liuchuanjian, xuchunjing, tian.qi1\}@huawei.com, } \small\texttt{xuchao@cis.pku.edu.cn}\\ 
}

	\maketitle

	\begin{abstract}
		Learning portable neural networks is very essential for computer vision for the purpose that pre-trained heavy deep models can be well applied on edge devices such as mobile phones and micro sensors. Most existing deep neural network compression and speed-up methods are very effective for training compact deep models, when we can directly access the training dataset. However, training data for the given deep network are often unavailable due to some practice problems (\eg privacy, legal  issue, and transmission), and the architecture of the given network are also unknown except some interfaces. To this end, we propose a novel framework for training efficient deep neural networks by exploiting generative adversarial networks (GANs). To be specific, the pre-trained teacher networks are regarded as a fixed discriminator and the generator is utilized for derivating training samples which can obtain the maximum response on the discriminator. Then, an efficient network with smaller model size and computational complexity is trained using the generated data and the teacher network, simultaneously. Efficient student networks learned using the proposed Data-Free Learning (DAFL) method achieve $92.22\%$ and $74.47\%$ accuracies using ResNet-18 without any training data on the CIFAR-10 and CIFAR-100 datasets, respectively. Meanwhile, our student network obtains an $80.56\%$ accuracy on the CelebA benchmark.
	
	\end{abstract}
	
	\section{Introduction}
	
	\begin{figure*}[t]
		\centering
		\includegraphics[width=0.85\linewidth,trim={0cm 0cm 0cm 0cm}]{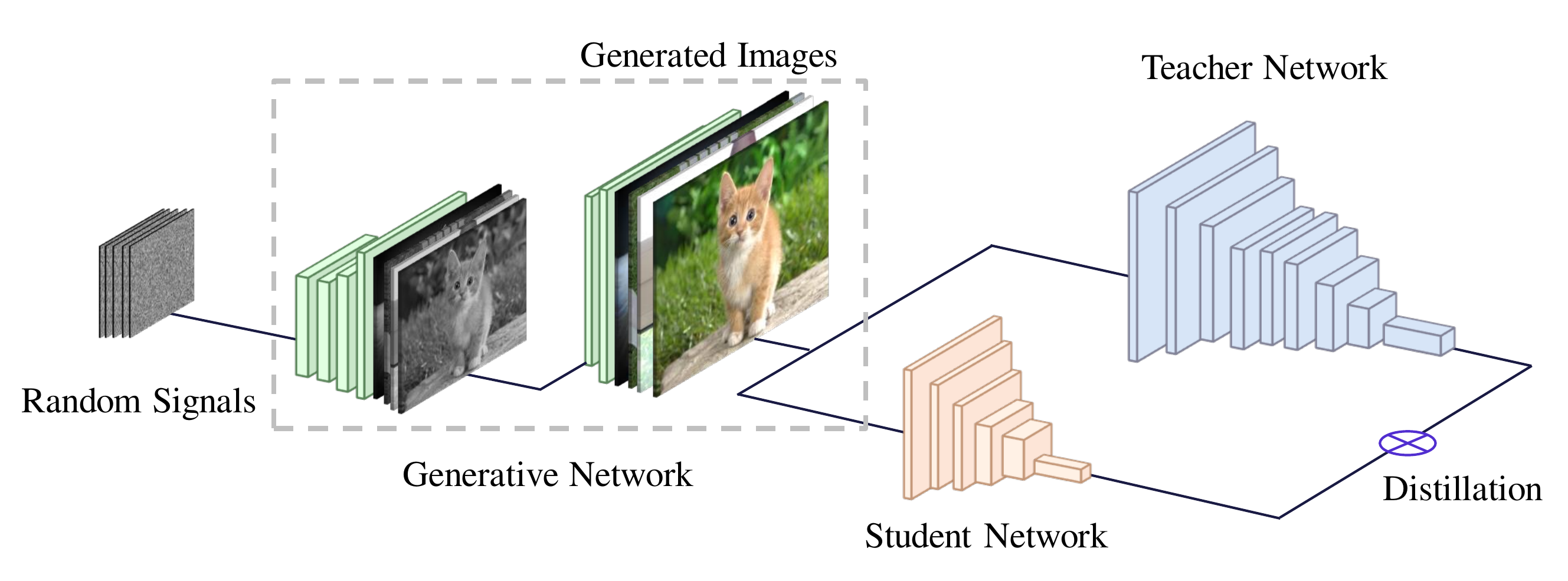}
		\caption{The diagram of the proposed method for learning efficient deep neural networks without the training dataset. The generator is trained for approximating images in the original training set by extracting useful information from the given network. Then, the portable student network can be effective learned by using generated images and the teacher network}
		\vspace{-0.5em}
		\label{fig1}
	\end{figure*}
	
	Deep convolutional neural networks (CNNs) have been successfully used in various computer vision applications such as image classification~\cite{VGG,krizhevsky2012imagenet}, object detection~\cite{ren2015faster} and semantic segmentation~\cite{long2015fully}. However, launching most of the widely used CNNs requires heavy computation and storage, which can only be used on PCs with modern GPU cards. For example, over $500$\emph{MB} of memory and over $10^{10}\times$ multiplications are demanded for processing one image using VGGNet~\cite{VGG}, which is almost impossible to be applied on edge devices such as autonomous cars and micro robots. Although these pre-trained CNNs have a number of parameters, Han~\etal~\cite{han2015deep} showed that discarding over $85\%$ of weights in a given neural network would not obviously damage its performance, which demonstrates that there is a significant redundancy in these CNNs.
	
	In order to compress and speed-up pre-trained heavy deep models, various effective approaches have been proposed recently. For example, Gong~\etal~\cite{gong2014compressing} utilized vector quantization approach to represent similar weights as cluster centers. Denton~\etal~\cite{SVD} exploited low-rank decomposition to process the weight matrices of fully-connected layers. Chen~\etal~\cite{Hash} proposed a hashing based method to encode parameters in CNNs. Han~\etal~\cite{han2015deep} employed pruning, quantization and Huffman coding to obtain a compact deep CNN with lower computational complexity. Hinton~\etal~\cite{hinton2015distilling} proposed the knowledge distillation approach, which distills the information of the pre-trained teacher network for learning a portable student network, \etc. 
	
	Although the above mentioned methods have made tremendous efforts on benchmark datasets and models, an important issue has not been widely noticed, \ie most existing network compression and speed-up algorithms have a strong assumption that training samples of the original network are available. However, the training dataset is routinely unknown in real-world applications due to privacy and transmission limitations. For instance, users do not want to let their photos leaked to others, and some of the training datasets are too huge to quickly upload to the cloud. In addition, parameters and architecture of pre-trained networks are also unknown sometimes except the input and output layers. Therefore, conventional methods cannot be directly used for learning portable deep models under these practice constrains. 
	
	Nevertheless, only a few works have been proposed for compressing deep models without training data. Lopes~\etal~\cite{Datafreekd} utilized the ``meta-data'' (\eg means and standard deviation of activations from each layer) recorded from the original training dataset, which is not provided for most well-trained CNNs. Srinivas and Babu~\cite{Datafreeprune} compressed the pre-trained network by merging similar neurons in fully-connected layers. However, the performance of compressed networks using these methods is much lower than that of the original network, due to they cannot effectively utilize the pre-trained neural networks. To address the aforementioned problem, we propose a novel framework for compressing deep neural networks without the original training dataset. To be specific, the given heavy neural network is regarded as a fixed discriminator. Then, a generative network is established for alternating the original training set by extracting information from the network during the adversarial procedure, which can be utlized for learning smaller networks with acceptable performance. The superiority of the proposed method is demonstrated through extensive experiments on benchmark datasets and models.
			
	Rest of this paper is organized as follows. Section~\ref{sec:related} investigates related works on CNN compression algorithms. Section~\ref{sec:method} proposes the data-free teacher-student paradigm by exploiting GAN. Section~\ref{sec:experi} illustrates experimental results of the proposed method on benchmark datasets and models and Section~\ref{sec:conclu} concludes the paper.

	\section{Related Works}\label{sec:related}
	
	Based on different assumptions and applications, existing portable network learning methods can be divided into two categories, \ie data-driven and data-free methods.
	
	\subsection{Data-Driven Network Compression}
	In order to learn efficient deep neural networks, a number of methods have been proposed to eliminate redundancy in pre-trained deep models. For example, Gong~\etal~\cite{gong2014compressing} employed the vector quantization scheme to represent similar weights in neural networks. Denton~\etal~\cite{SVD} exploited the singular value decomposition (SVD) approach to decompose weight matrices of fully-connected layers. Han~\etal~\cite{han2015deep} proposed the pruning approach for removing subtle weights in pre-trained neural networks. Wang~\etal~\cite{wang2016cnnpack} further introduced the discrete cosine transform (DCT) bases and converted convolution filters into the frequency domain to achieve higher compression and speed-up ratios. Yang~\etal~\cite{yang2019legonet} used a set of Lego filters to build efficient CNNs.
	
	Besides eliminating redundant weights or filters, Hinton~\etal~\cite{hinton2015distilling} proposed a knowledge distillation (KD) paradigm for transferring useful information from a given teacher network to a portable student network.  Yim~\etal~\cite{yim2017gift} introduced the FSP (Flow of Solution Procedure) matrix to inherit the relationship between features
	from two layers. Li~\etal~\cite{li2017mimicking} further presented a feature mimic framework to train efficient convolutional networks for objective detection. 
	In addition, Rastegari~\etal~\cite{rastegari2016xnor} and Courbariaux~\etal~\cite{courbariaux2016binarized} explored binarized neural networks to achieve considerable compression and speed-up ratios, which weights are -1/1 or -1/0/1, \etc.
	
	Although the above mentioned algorithms obtained promising results on most of benchmark datasets and deep models, they cannot be effectively launched without the original training dataset. In practice, the training dataset could be unavailable for some reasons, \eg transmission limitations and privacy. Therefore, it is necessary to study the data-free approach for compressing neural networks.
	
	\subsection{Data-Free Network Compression}
	There are only a few methods that are proposed for compressing deep neural networks without the original training dataset. Srinivas and Babu~\cite{Datafreeprune} proposed to directly merge similar neurons in fully-connected layers, which cannot be applied on convolutional layers and networks which detail architectures and parameters information are unknown. In addition, Lopes~\etal~\cite{Datafreekd} attempted to reconstruct the original data from ``meta-data'' and utilize the knowledge distillation scheme to learn a smaller network.
		
	Since the fine-tuning procedure cannot be accurately conducted without the original training dataset, performance of compressed methods by existing algorithms is worse than that of baseline models. Therefore, an effective data-free approach for learning efficient CNNs with comparable performance is highly required.
	
	\section{Data-free Student Network learning}\label{sec:method}

	In this section, we will propose a novel data-free framework for compressing deep neural networks by embedding a generator network into the teacher-student learning paradigm.
	
	\subsection{Teacher-Student Interactions}
	As mentioned above, the original training dataset is not usually provided by customers for various concerns. In addition, parameters and detailed architecture information could also be unavailable sometimes. Thus, we propose to utilized the teacher-student learning paradigm for learning portable CNNs.
		
	Knowledge Distillation (KD)~\cite{hinton2015distilling} is a widely used approach to transfer the output information from a heavy network to a smaller network for achieving higher performance, which does not utilize parameters and the architecture of the given network. Although the given deep models may only be provided with limited interfaces (\eg input and output interfaces), we can transfer the knowledge to inherit the useful information from the teacher networks. Let $\mathcal{N}_{T}$ and $\mathcal{N}_{S}$ denote the original pre-trained convolutional neural network (teacher network) and the desired portable network (student network), the student network can be optimized using the following loss function based on knowledge distillation:
	\begin{equation}
	\label{Fcn7}
	\mathcal{L}_{KD} = \frac{1}{n}\sum\limits_{i} \mathcal{H}_{cross}(\mbox{y}_S^i,\mbox{y}_T^i).
	\end{equation}
	where $\mathcal{H}_{cross}$ is the cross-entropy loss, $\mbox{y}_T^i=\mathcal{N}_T(\mbox{x}^i)$ and $\mbox{y}_S^i=\mathcal{N}_S(\mbox{x}^i)$ are the outputs of the teacher network $\mathcal{N}_T$ and student network $\mathcal{N}_S$, respectively. Therefore, utilizing the knowledge transfer technique, a portable network can be optimized without the specific architecture of the given network.
	
	\subsection{GAN for Generating Training Samples}
	In order to learn portable network without original data, we exploit GAN to generate training samples utilizing the available information of the given network.
	
	Generative adversarial networks (GANs) have been widely applied for generating samples. GANs consist of a generator $G$ and a discriminator $D$. $G$ is expected to generate desired data while $D$ is trained to identify the differences between real images and those produced by the generator. To be specific, given an input noise vector $z$, $G$ maps $z$ to the desired data $\mbox{x}$, \ie $G:\mbox{z}\rightarrow \mbox{x}$. On the other hand, the goal of $D$ is to distinguish the real data from synthetic data $G(\mbox{z})$. For an aribitrary vanilla GAN, the objective function can be formulated as
	\begin{equation}
	\begin{aligned}
	\mathcal{L}_{GAN} = &\mathbb{E}_{y\sim p_{data}(y)}[\log D(y)] \\
	+& \mathbb{E}_{z\sim p_{z}(z)}[\log (1-D(G(z)))].
	\end{aligned}
	\label{Fcn1}
	\end{equation}
	In the adversarial procedure, the generator is continuously upgraded according to the training error produced by $D$. The optimal $G$ is obtained by optimizing the following problem
	\begin{equation}
	G^*=\mbox{arg}\min\limits_{G} \mathbb{E}_{z\sim p_{z}(z)}[\log (1-D^{*}(G(z)))],
	\label{Fcn2}
	\end{equation}
	where $ D^{*}$ is the optimal discriminator. Adversarial learning techniques can be naturally employed to synthesize training data. However according to Eq. (\ref{Fcn1}), the discriminator requires real images for training. In the absence of training data, it is thus impossible to train the discriminator as vanilla GANs. 
	
	Recent works~\cite{radford2015unsupervised} have proved that the discriminator $D$ can learn the hierarchy of representations from samples, which encourages the generalization of $D$ in other tasks like image classification. Odena~\cite{odena2016semi} further suggested that the tasks of discrimination and classification can improve each other. Instead of training a new discriminator as vanilla GANs, the given deep neural network can extract semantic features from images as well, since it has already been well trained on large-scale datasets. Hence, we propose to regard this given deep neural network (\eg ResNet-50~\cite{he2016deep}) as a fixed discriminator. Therefore, $G$ can be optimized directly without training $D$ together, \ie the parameters of original network $D$ are fixed during training $G$. In addition, the output of the discriminator is a probability indicating whether an input image is real or fake in vanilla GANs. However, given the teacher deep neural network as the discriminator, the output is to classify images to different concept sets, instead of indicating the reality of images. The loss function in vanilla GANs is therefore inapplicable for approximating the original training set. Thus, we conduct thorough analysis on real images and their responses on this teacher network. Several new loss functions will be devised to reflect our observations. 
	
	On the image classification task, the teacher deep neural network adopts the cross entropy loss in the training stage, which enforces the outputs to be close to ground-truth labels of inputs. Specifically for multi-class classification, the outputs are encouraged to be one-hot vectors, where only one entry is 1 and all the others are 0s. Denote the generator and the teacher network as $G$ and $\mathcal{N}_T$, respectively. Given a set of random vector $\{\mbox{z}^1,\mbox{z}^2,\cdots,\mbox{z}^n\}$, images generated from these vectors are $\{\mbox{x}^1,\mbox{x}^2,\cdots,\mbox{x}^n\}$, where $\mbox{x}^i=G(\mbox{z}^i)$. Inputting these images into the teacher network, we can obtain the outputs $\{\mbox{y}^1_T,\mbox{y}^2_T,\cdots,\mbox{y}^n_T\}$ with $\mbox{y}_T^i = \mathcal{N}_T(\mbox{x}^i) $. The predicted labels $\{\mbox{t}^1,\mbox{t}^2,\cdots,\mbox{t}^n\}$ are then calculated by $\mbox{t}^i=\mbox{arg}\max\limits_{j} (\mbox{y}_T^i)_j$. If images generated by $G$ follow the same distribution as that of the training data of the teacher network, they should also have similar outputs as the training data. We thus introduce the one-hot loss, which encourages the outputs of generated images by the teacher network to be close to one-hot like vectors. By taking $\{\mbox{t}^1,\mbox{t}^2,\cdots,\mbox{t}^n\}$ as pseudo ground-truth labels, we formulate the one-hot loss function as 
	\begin{equation}
	\mathcal{L}_{oh} = \frac{1}{n}\sum\limits_{i}\mathcal{H}_{cross}(\mbox{y}_T^i,\mbox{t}^{i}) ,
	\label{Fcn3}
	\end{equation}
	where $\mathcal{H}_{cross}$ is the cross-entropy loss function. By introducing the one-hot loss, we expect that a generated image can be classified into one particular category concerned by the teacher network with a higher probability. In other words, we pursue synthetic images that are exclusively compatible with the teacher network, rather than general real images for any scenario. 
	
	Besides predicted class labels by DNNs, intermediate features extracted by convolution layers are also important representations of input images. A large number of works have investigated the interpretability of deep neural networks ~\cite{zeiler2014visualizing,selvaraju2017grad,dong2017towards}. Features extracted by convolution filters are supposed to contain valuable information about the input images. In particular, Zhang~\etal~\cite{zhang2018interpretable} assigned each filter in a higher convolution layer with a part of object, which demonstrates that each filter stands for different semantics. We denote features of $\mbox{x}^i$ extracted by the teacher network as $f^i_{T}$, which corresponds to the output before the fully-connected layer.  Since filters in the teacher DNNs have been trained to extract intrinsic patterns in training data, feature maps tend to receive higher activation value if input images are real rather than some random vectors. Hence, we define an activation loss function as:
	\begin{equation}
	\mathcal{L}_{a} = -\frac{1}{n}\sum\limits_{i}\Vert f^i_{T}\Vert _1,
	\label{Fcn4}
	\end{equation}
	where $\Vert \cdot \Vert _1$ is the conventional $l_1$ norm.

	\begin{algorithm}[t] 
		\caption{DAFL for learning portable student networks.} 
		\label{alg1} 
		\begin{algorithmic}[1] 
			\Require 
			A given teacher network $\mathcal{N}_T$, parameters of different objects: $\alpha$ and $\beta$. 
			\State Initialize the generator $G$, the student network $\mathcal{N}_S$ with fewer memory usage and computational complexity;
			\Repeat
			\State \textbf{Module 1: Training the Generator.}
			\State Randomly generate a batch of vector: $\{\mbox{z}^i\}^n_{i=1}$;
			\State Generate the training samples: $ x \leftarrow G(\mbox{z})$;
			\State Employ the teacher network on the mini-batch: \\\quad\quad\quad\quad\quad\quad$ [\mbox{y}_T,\mbox{t},f_T] \leftarrow \mathcal{N}_T(\mbox{x})$;	
			\State Calculate the loss function $\mathcal{L}_{Total}$ (Fcn.\ref{Fcn6}):
			\State Update weights in $G$ using back-propagation; 
			\State \textbf{Module 2: Training the student network.}
			\State Randomly generate a batch of vector $\{\mbox{z}^i\}^n_{i=1}$;
			\State Utlize the generator on the mini-batch: $ x \leftarrow G(\mbox{z})$;
			\State Employ the teacher network and the student network on the mini-batch simultaneously: \\\quad\quad\quad\quad\quad$ \mbox{y}_S \leftarrow \mathcal{N}_S(\mbox{x})$, $ \mbox{y}_T \leftarrow \mathcal{N}_T(\mbox{x})$;
			\State Calculate the knowledge distillation loss:\\\quad\quad\quad\quad\quad $\mathcal{L}_{KD} \leftarrow \frac{1}{n}\sum\limits_{i} \mathcal{H}(\mbox{y}_S^i,\mbox{y}_T^i)$;
			\State Update weights in $\mathcal{N}_S$ according to the gradient;
			\Until convergence
			\Ensure 
			The student network $\mathcal{N}_S$.
		\end{algorithmic} 
	\end{algorithm}
	
		\begin{table*}[t]
		\small
		\begin{center}
			\caption{Classification result on the MNIST dataset.}
			\label{table:cls}
			\begin{tabular}{|c|c|c|c|c|c|c|c|}
				\hline
				\multirow{2}{*}{\textbf{Algorithm}} & \multirow{2}{*}{\textbf{Required data}} &\multicolumn{3}{|c|}{\textbf{LeNet-5~\cite{LeNet}}}&\multicolumn{3}{|c|}{\textbf{HintonNet~\cite{hinton2015distilling}}}\\
				\cline{3-8} &&\textbf{Accuracy} &\textbf{FLOPs} &\textbf{\#params}& \textbf{Accuracy} &\textbf{FLOPs} &\textbf{\#params} \\
				\hline
				\hline
				Teacher &Original data& 98.91\% &$\sim$436K &$\sim$62K&98.39\%& $\sim$2.39M & $\sim$2.4M\\
				\hline
				Standard back-propagation & Original data& 98.65\% &$\sim$144K& $\sim$16K&98.11\% &$\sim$1.28M&$\sim$ 1.28M\\
				\hline
				{Knowledge Distillation\small{~\cite{hinton2015distilling}}}& Original data&98.91\%&$\sim$144K& $\sim$16K&98.39\%&$\sim$1.28M&$\sim$ 1.28M\\
				\hline
				\hline
				Normal distribution & No data & 88.01\%&$\sim$144K& $\sim$16K&87.58\%&$\sim$1.28M&$\sim$ 1.28M\\
				\hline
				Alternative data & USPS dataset & 94.56\%&$\sim$144K& $\sim$16K&93.99\%&$\sim$1.28M&$\sim$ 1.28M\\
				\hline
				Meta data~\cite{Datafreekd} & Meta data & 92.47\%&$\sim$144K& $\sim$16K&91.24\%&$\sim$1.28M&$\sim$ 1.28M\\
				\hline
				Data-Free Learning (DAFL) & No data &98.20\%&$\sim$144K& $\sim$16K& 97.91\%&$\sim$1.28M&$\sim$ 1.28M\\
				\hline
			\end{tabular}
		\end{center}
		\vspace{-0.5em}
	\end{table*}
	
	Moreover, to ease the training procedure of a deep neural network, the number of training examples in each category is usually balanced, \eg there are 6,000 images in each class in the MNIST dataset. We employ the information entropy loss to measure the class balance of generated images. Specifically, given a probability vector $\mbox{p} = (p_1,p_2,\cdots,p_k)$, the information entropy, which measures the degree of confusion, of $\mbox{p}$ is calculated as $\mathcal{H}_{info}(\mbox{p}) = -\frac{1}{k}\sum\limits_{i}  p_i \log(p_i)$. The value of $\mathcal{H}_{info}(\mbox{p})$ indicates the amount of information that $\mbox{p}$ owns, which will take the maximum when all variables equal to $\frac{1}{k}$. Given a set of output vectors $\{\mbox{y}_T^1,\mbox{y}_T^2,\cdots,\mbox{y}^n_T\}$, where $\mbox{y}^i_T = \mathcal{N}_T(\mbox{x}^i)$, the frequency distribution of generated images for every class is $\frac{1}{n}\sum\limits_{i} \mbox{y}^i_T$. The information entropy loss of generated images is therefore defined as
	\begin{equation}
	\mathcal{L}_{ie} = -\mathcal{H}_{info}( \frac{1}{n}\sum\limits_{i} \mbox{y}^i_T).
	\label{Fcn5}
	\end{equation}
	When the loss takes the minimum, every element in vector $\frac{1}{n}\sum\limits_{i} \mbox{y}^i_S$ would equal to $\frac{1}{k}$, which implies that $G$ could generate images of each category with roughly the same probability. Therefore, minimizing the information entropy of generated images can lead to a balanced set of synthetic images.
	
	By combining the aforementioned three loss functions, we obtain the final objective function
	\begin{equation}
	\label{Fcn6}
	\mathcal{L}_{Total} = \mathcal{L}_{oh} + \alpha\mathcal{L}_{a} + \beta\mathcal{L}_{ie},
	\end{equation}
	where $\alpha$ and $\beta$ are hyper parameters for balancing three different terms. By minimizing the above function, the optimal generator can synthesize images that have the similar distribution as that of the training data previously used for training the teacher network (\ie the discriminator network). 
	
	It is noted that some previous works~\cite{simonyan2013deep,mahendran2015understanding} could synthesize images by optimizing the input of the neural network using back-propagation. But it is difficult to generate abundant images for the subsequent student network training, for each synthetic image leads to an independent optimization problem solved by back-propagation. In contrast, the proposed method can imitate the distribution of training data directly, which is more flexible and efficient to generate new images.

	
	\subsection{Optimization}

	\begin{table*}[t]
		\begin{center}
			\caption{Effectiveness of different components of the proposed data-free learning method.}
			\label{table:abl}
			\begin{tabular}{|c|c|c|c|c|c|c|c|c|}
				\hline
				One-hot loss & & \Checkmark & & &\Checkmark& \Checkmark& &\Checkmark\\
				\hline
				Information entropy loss &  & & \Checkmark &&& \Checkmark&\Checkmark &\Checkmark\\
				\hline
				Feature maps activation loss &  & & &\Checkmark &\Checkmark& &\Checkmark&\Checkmark\\
				\hline
				\textbf{Top 1 accuracy} & 88.01\% &78.77\%& 88.14\% &15.95\%& 42.07\%& 97.25\%& 95.53\%&\textbf{98.20\%}  \\
				\hline
			\end{tabular}
		\end{center}
		\vspace{-0.5em}
	\end{table*}
	
	The learning procedure of our algorithm can be divided into two stages of training. First, we regard the well-trained teacher network as a fixed discriminator. Using the loss function $\mathcal{L}_{Total}$ in Eq.~\ref{Fcn6}, we optimize a generator $G$ to generate images that follow the similar distribution as that of the original training images for the teacher network. Second, we utilize the knowledge distillation approach to directly transfer knowledge from the teacher network to the student network. The student network with fewer parameters is then optimized using the KD loss $\mathcal{L}_{KD}$ in Eq.~\ref{Fcn7}. The diagram of the proposed method is shown in Figure~\ref{fig1}.
	
	We use stochastic gradient descent (SGD) method to optimize the image generator $G$ and the student network $\mathcal{N}_{S}$. In the training of $G$, the first term of $\mathcal{L}_{Total}$  is the cross entropy loss, which can be trained traditionally. The second term $\mathcal{L}_{a}$ in Eq.~\ref{Fcn6} is exactly a linear operation, and the gradient of $\mathcal{L}_{a}$ with respect to $f^i_{T}$ can be easily calculated as:
	\begin{equation}
	\frac{\partial\mathcal{L}_{a}}{\partial f_{T}^i} = -\frac{1}{n} \mbox{sgn}(f^i_T),
	\end{equation}
	where $\mbox{sgn}(\cdot)$ denotes sign function. Parameters $W_{G}$ in $G$ will be updated by:
	\begin{equation}
	\frac{\partial \mathcal{L}_{a}}{\partial W_{G}} = \sum\limits_{i}\frac{\partial \mathcal{L}_{a}}{\partial f_{T}^i} \cdot\frac{\partial f_{T}^i}{\partial W_{G}},
	\end{equation}
	where $\frac{\partial f_{T}^i}{\partial W_{G}}$ is the gradient of the feature $ f_{T}^i$. 
	The gradient of the final term $\mathcal{L}_{ie}$ with respect to $\mbox{y}_T^i$ can be easily calculated as:
	\begin{equation}
	\frac{\partial\mathcal{L}_{ie}}{\partial \mbox{y}_T^i} = -\frac{1}{n} \mbox{y}^i[\log(\frac{1}{n}\sum\limits_{j} \mbox{y}_T^j)+\mathbf{1}],
	\end{equation}
	where $\mathbf{1}$ denotes $n$-dimensional vector with all values as $1$. Parameters in $G$ will be additionally updated by:
	\begin{equation}
	\frac{\partial \mathcal{L}_{ie}}{\partial W_{G}} = \sum\limits_{i}\frac{\partial \mathcal{L}_{ie}}{\partial \mbox{y}_T^i} \cdot\frac{\partial \mbox{y}_T^i}{\partial W_{G}}.
	\end{equation}
	Detailed procedures of the proposed Data-Free Learning (DAFL) scheme for learning efficient student neural networks is summarized in Algorithm 1.

	\section{Experiments}\label{sec:experi}
	
	In this section, we will demonstrate the effectiveness of our proposed data-free knowledge distillation method and conduct massive ablation experiments to have an explicit understanding of each component in the proposed method.
	
	\subsection{Experiments on MNIST}\label{sec:cls}

	We first implement experiments on the MNIST dataset, which is composed of $28\times28$ pixel images from 10 categories (from 0 to 9). The whole dataset consists of 60,000 training images and 10,000 testing images. For choosing hyper-parameters of the proposed methods, we take 10,000 images as a validation set from training images. Then, we train models on the full 60,000 images to obtain the ultimate network.
	
	To make a fair comparison, we follow the setting in ~\cite{Datafreekd}. Two architectures are used for investigating the performance of proposed method, \ie a convolution-based architecture and a network consists of fully-connect layers. For convolution models, we use LeNet-5~\cite{LeNet} as the teacher model and LeNet-5-HALF (a modified version with half the number of channels per layer) as the student model. For the second architecture, the teacher network consists of two hidden layers of 1,200 units
	(Hinton-784-1200-1200-10)~\cite{hinton2015distilling} and student network consists of two hidden layers of 800 units (Hinton-784-800-800-10). The student networks have significantly fewer parameters than teacher networks. 
	For our method, $\alpha$ and $\beta$ in Fcn.\ref{Fcn6} are 0.1 and 5, respectively, and are tuned on the validation set. The generator was trained for 200 epochs using Adam. We use a deep convolutional generator\footnote{{https://github.com/eriklindernoren/PyTorch-GAN/blob/master/implementations/dcgan/dcgan.py}} following~\cite{radford2015unsupervised} and add a batch normalization at the end of the generator to smooth the sample values. 
	
	Table~\ref{table:cls} reports the results of different methods on the MNIST datasets. On LeNet-5 models, the teacher network achieves a $98.91\%$ accuracy while the student network using the standard back-propagation achieves a $98.65\%$ accuracy, respectively. Knowledge distillation improved the accuracy of student network to $98.91\%$. These methods use the original data to train the student network. We then train a student network exploiting the proposed method to evaluate the effectiveness of the synthetic data.
	
	We first use the data randomly generated from normal distribution to training the student network. By utilizing the knowledge distillation, the student network achieves only an $88.01\%$ accuracy. In addition, we further use another handwritten digits dataset, namely USPS~\cite{usps}, to conduct the same experiment for training the student network. Although images in two datasets have similar properties, the student network learned using USPS can only obtain a 94.56\% accuracy on the MNIST dataset, which demonstrates that it is extremely hard to find an alternative to the original training dataset. To this end,
	Lopes~\etal~\cite{Datafreekd} using the ``meta data'', which is the activation record of original data, to reconstruct the dataset and achieved only a 92.47\% accuracy. Noted that the upper bound of the accuracy of student network is 98.65\%, which could be achieved only if we could find a dataset whose distribution is same as the original dataset (\ie MNIST dataset). The proposed method utilizing generative adversarial networks achieved a 98.20\% accuracy, which is much close to this upper bound. Also, the accuracy of student network using the proposed algorithm is superior to these using other data (normal distribution, USPS dataset and reconstructed dataset using ``meta data''), which suggest that our method could imitate the distribution of training dataset better.
	
	On the fully-connected models, the classification accuracies of teacher and student network are $98.39\%$ and $98.11\%$, respectively. Knowledge Distillation brought the performance of student network by transferring information from teacher network to $98.39\%$. However, in the absence of training data, the result became unacceptable. Randomly generated noise only achieves $87.58\%$ accuracy and ``meta data''~\cite{Datafreekd} achieves a higher accuracy of $91.24\%$. Using USPS dataset as alternatives achieves an accuracy of 93.99\%. The proposed method results in the highest performance of $97.91\%$ among all methods without the original data, which demonstrates the effectiveness of the generator. 
	
	\begin{table*}[t]
		\begin{center}
			\caption{Classification result on the CIFAR dataset.}
			\label{table:cifar}
			\begin{tabular}{|c|c|c|c|c|c|}
				\hline
				\textbf{Algorithm} &\textbf{Required data}&\textbf{FLOPS} &\textbf{\#params}&\textbf{CIFAR-10}&\textbf{CIFAR-100}  \\
				\hline
				\hline
				Teacher& Original data&$\sim$1.16G &$\sim$21M&95.58\% & 77.84\%\\
				\hline
				Standard back-propagation & Original data&$\sim$557M  & $\sim$11M& 93.92\% &  76.53\%\\
				\hline
				{Knowledge Distillation{~\cite{hinton2015distilling}}}& Original data&$\sim$557M  & $\sim$11M&94.34\%& 76.87\%\\
				\hline
				\hline
				Normal distribution &No data&$\sim$557M  & $\sim$11M& 14.89\% & 1.44\%\\
				\hline
				Alternative data &Similar data&$\sim$557M  & $\sim$11M& 90.65\% & 69.88\%\\
				\hline
				Data-Free Learning (DAFL) & No data&$\sim$557M  & $\sim$11M&92.22\% & 74.47\%\\
				\hline
			\end{tabular}
		\end{center}
		\vspace{-0.5em}
	\end{table*}

	\subsection{Ablation Experiments}\label{sec:abl}

	In the above sections, we have tested and verified the effectiveness of the proposed generative method for student network learning without training data. However, there are a number of components, \ie three terms in Eq.~\ref{Fcn6}, when optimizing the generator. We further conduct the ablation experiments for an explicit understanding and analysis.
	
	The ablation experiment is also conducted on the MNIST dataset. We used the LeNet-5 as a teacher network and LeNet-5-HALF as a student network. The training settings are same as those in Section~\ref{sec:cls}. Table~\ref{table:abl} reports the results of various design components. Using randomly generated samples, \ie the generator $G$ is not trained, the student network achieves an 88.01\% accuracy. However, by utilizing one-hot loss and feature map activation loss or one of them, the generated samples are unbalanced, which results in the poor performance of the student networks. Only introducing information entropy loss, the student network achieves an 88.14\% accuracy since the samples do not contain enough useful information. When combining $\mathcal{L}_{oh}$ or $\mathcal{L}_{a}$ with $\mathcal{L}_{ie}$, the student network achieves higher performance of 97.25\% and 95.53\%, respectively. Moreover, the accuracy of student network is 98.20\% when using all these loss functions, which achieves the best performance. It is worth noticing that the combination of one-hot loss and information entropy is essential for training the generator, which is also utilized in some previous works~\cite{springenberg2015unsupervised,jain2017subic}.
	
	The ablation experiments suggest that each component of the loss function of $G$ is meaningful. By applying the proposed method, $G$ can generate balanced samples from different classes with a similar distribution as that in the original dataset, which is effective for the training of the student network.
	
	\subsection{Visualization Results}\label{sec:vis}
	
	After investigating the effectiveness of the proposed method, we further conduct visualization experiments on the MNIST dataset. There are 10 categories of handwritten digits from 0 to 9 in the MNIST dataset. The settings are same as that in Section~\ref{sec:cls}. 
	
	\begin{figure}[h]
		\vspace{-0.5em}
		\centering
		\includegraphics[scale=1.3]{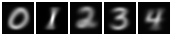}\\
		\includegraphics[scale=1.3]{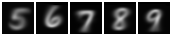}\\
		(a) Averaged images on the MNIST dataset.\\
		\includegraphics[scale=1.3]{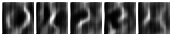}\\
		\includegraphics[scale=1.3]{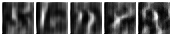}\\
		(b) Averaged images on the generated dataset.
		\vspace{0.3em}
		\caption{Visualization of averaged image in each category (from 0 to 9) on the MNIST dataset.
		}
		\vspace{-1.5em}
		\label{fig:image}
	\end{figure}
	
	Figure~\ref{fig:image} shows the visualization results of averaged images. Noted that the generated images are unlabeled, their classes are defined by the prediction of the teacher network. By exploiting the information of the given network as much as possible, we design loss function for the generator. Figure~\ref{fig:image} (b) shows the mean of images of each class. Although no real image is provided, the generated images have similar patterns with the training images, which indicates that the generator can somehow learn the data distribution.

\begin{figure}[h]
		\centering
		\includegraphics[width=0.6\linewidth]{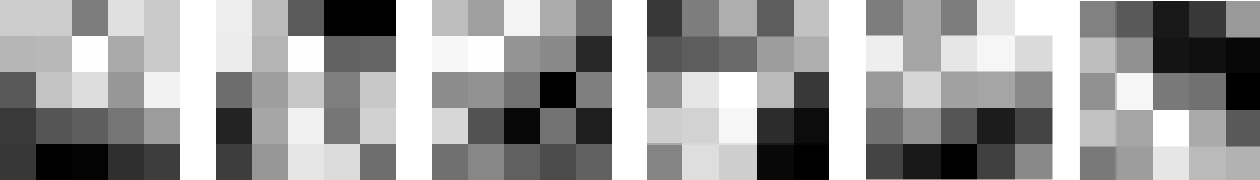}\\
		(a) Teacher filters. \\
		\vspace{1em}
		\includegraphics[width=0.6\linewidth]{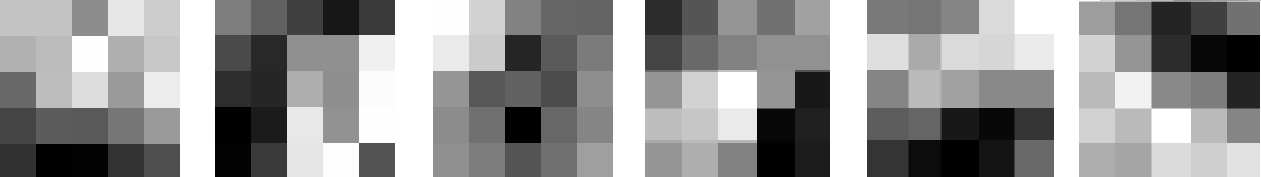}\\
		(b) Student filters.
		\vspace{0.5em}
		\caption{Visualization of filters in the first convolutional layer learned on the MNIST dataset. The top line shows filters trained using the original training dataset, and the bottom line shows filters obtained using samples generated by the proposed method.}
		\label{fig:filter}
		\vspace{-1.5em}
	\end{figure}

	\textbf{Filter visualization.}
	Moreover, we visualize the filters of the LeNet-5 teacher network and student network in Figure~\ref{fig:filter}. Though the student network is trained without real-world data, filters of the student network learned by the proposed method (see Figure~\ref{fig:filter} (b)) are still similar to those of the teacher network (see Figure~\ref{fig:filter} (a)). The visualization experiments further demonstrate that the generator can produce images that have similar patterns as the original images, and by utilizing generated samples, the student network could acquire valuable knowledge from the teacher network. 
	
	\subsection{Experiments on CIFAR}
	
	To further evaluate the effectiveness of our method, we conduct experiments on the CIFAR dataset. 
	We used a ResNet-34 as the teacher network and ResNet-18 as the student network\footnote{https://github.com/kuangliu/pytorch-cifar}, which is complex and advanced for further investigating the effectiveness of the proposed method. These networks are optimized using Nesterov Accelerated Gradient (NAG) and the weight decay and the momentum are set as $5\times10^{-4}$ and 0.9, respectively. We train the networks for 200 epochs and the initial learning rate is set as 0.1 and divided by 10 at 80 and 120 epochs, respectively. Random flipping, random crop and zero padding are used for data augmentation as suggested in~\cite{he2016deep}. $G$ and the student networks of the proposed method are trained for 2,000 epochs and the other settings are same as those in MNIST experiments.
	
	Table~\ref{table:cifar} reports the classification results on the CIFAR-10 and CIFAR-100 datasets. The teacher network achieves a 95.58\% accuracy in CIFAR-10. The student network using knowledge distillation achieves a 94.34\% accuracy, which is slightly higher than that of standard BP (93.92\%).

	We then explore to optimize the student network without true data. Since the CIFAR dataset is more complex than MNIST, it is impossible to optimize a student network using randomly generated data which follows the normal distribution. Therefore, we then regard the MNIST dataset without labels as an alternative data to train the student network using the knowledge distillation. The student network only achieves a 28.29\% accuracy on the CIFAR-10 dataset. Moreover, we train the student network using the CIFAR-100 dataset, which has considerable overlaps with the original CIFAR-10 dataset, but this network only achieves a 90.65\% accuracy, which is obviously lower than that of the teacher model. In contrast, the student network trained utilizing the proposed method achieved a 92.22\% accuracy with only synthetic data.
	
	Besides CIFAR-10, we further verify the capability of the proposed method on the CIFAR-100 dataset, which has 100 categories and 600 images per class. Therefore, the dimensionality of the input random vectors for the generator in our method is increased to 1,000. The accuracy of the teacher network is 77.84\% and that of the student network is only 76.53\%, respectively.  Using normal distribution data, MNIST, and CIFAR-10 to train the student network cannot obtain promising results, as shown in Table~\ref{table:cifar}. In contrast, the student network learned by exploiting the proposed method obtained a 74.47\% accuracy without any real-world training data.

	\subsection{Experiments on CelebA}

	Besides the CIFAR dataset, we conduct our experiments on the CelebA dataset, which contains 202,599 face images of pixel $224\times224$. To evaluate our approach fairly, we used AlexNet~\cite{krizhevsky2012imagenet} to classify the most balanced attribute in CelebA~\cite{liu2015deep} following the settings in~\cite{Datafreekd}. The student network is AlexNet-Half, which number of filters is half of AlexNet. The original teacher network has about 57M parameters while the student network has only about 40M parameters. The networks is optimized for 100 epochs using Adam with a learning rate of $10^{-4}$. We use an alternative model of DCGAN~\cite{radford2015unsupervised} to generate color images of $224\times224$. The hyper-parameters of the proposed method are same as those in MNIST and CIFAR experiments and $G$.

	Table \ref{table:CelebA} reported the classification results of student networks on the CelebA dataset by exploiting the proposed method and state-of-the-art learning methods. The teacher network achieves an 81.59\% accuracy and the student network using the standard BP achieves an 80.82\% accuracy, respectively. Lopes~\etal~\cite{Datafreekd} achieves only a 77.56\% accuracy rate using the ``meta data''. The accuracy of the student network trained using the proposed method is 80.03\%, which is comparable with that of the teacher network.
	
	\begin{table}[!h]
	\begin{center}
		\caption{Classification result on the CelebA dataset.}
		\label{table:CelebA}
		\begin{tabular}{|c|c|c|}
			\hline
			\textbf{Algorithm} &\textbf{FLOPS}& \textbf{Accuracy} \\
			\hline
			\hline
			Teacher&$\sim$711M& 81.59\%\\
			\hline
			Standard back-propagation&$\sim$222M & 80.82\%\\
			\hline
			{Knowledge Distillation\small{~\cite{hinton2015distilling}}}&$\sim$222M&81.35\%\\
			\hline
			\hline
			Meta data~\cite{Datafreekd} &$\sim$222M& 77.56\% \\
			\hline
			Data-Free Learning (DAFL) &$\sim$222M&80.03\% \\
			\hline
		\end{tabular}
	\end{center}
	\vspace{-1.5em}
    \end{table}
	
	\subsection{Extended Experiments}

	Massive experiments are conducted on several benchmarks to verify the performance of the DAFL method for learning student networks using generated images. Wherein, architectures of used student networks are more portable than those of teacher networks. To investigate the difference between original training images and generated images, we use these generated images to train networks of the same architectures as those of teacher networks using the proposed methods. The results are reported in Table~\ref{table:variouscls}.
	
	It can be found in Table~\ref{table:variouscls} that LeNet-5 and HintonNet on the MNIST dataset achieve a 98.91\% accuracy and a 98.39\% accuracy, respectively. In contrast, accuracies of student networks trained from scratch with same architectures are 98.47\% and 98.08\%, respectively, which are very close to those of teacher networks. In addition, student networks on the CIFAR-10 and the CIFAR-100 datasets also obtain similar results to those of teacher networks. These results demonstrate that the proposed method can effectively approximate the original training dataset by extracting information from teacher networks. If the network architectures are given, we can even replicate the teacher networks and achieve similar accuracies.

	\begin{table}[h]
	\begin{center}
		\vspace{-0.5em}
		\caption{Classification results on various datasets.}
		\label{table:variouscls}
		\begin{tabular}{|c|c|c|c|c|}
			\hline
			\multirow{2}{*}{\textbf{Dataset}}&\multirow{2}{*}{\textbf{Model}}&\multicolumn{2}{|c|}{\textbf{Accuracy}}\\
			\cline{3-4}
			&&\textbf{Teacher}&\textbf{Student}\\
			\hline
			MNIST&LeNet-5~\cite{LeNet}&98.91\%&98.47\%\\
			\hline
			MNIST&HintonNet~\cite{hinton2015distilling}&98.39\%&98.08\%\\
			\hline
			CIFAR-10&ResNet-34~\cite{he2016deep}&95.58\%&93.21\%\\
			\hline
			CIFAR-100&ResNet-34~\cite{he2016deep}&77.84\%&75.32\%\\
			\hline
			CelebA&AlexNet~\cite{krizhevsky2012imagenet}&81.59\%&80.56\%\\
			\hline
		\end{tabular}
	\end{center}
	\vspace{-1.5em}
    \end{table}

	\section{Conclusion}\label{sec:conclu}

	Conventional methods require the original training dataset for fine-tuning the compressed deep neural networks with an acceptable accuracy. However, the training set and detailed architecture information of the given deep network are routinely unavailable due to some privacy and transmission limitations. In this paper, we present a novel framework to train a generator for approximating the original dataset without the training data. Then, a portable networks can be learned effectively through the knowledge distillation scheme. Experiments on benchmark datasets demonstrate that the proposed method DAFL method is able to learn portable deep neural networks without any training data.
	
	\section*{Acknowledgement}
	
	This work is supported by National Natural Science Foundation of China under Grant No. 61876007, 61872012 and
	Australian Research Council Project DE-180101438.
	
	{
		\small
		\bibliographystyle{ieee_fullname}
		\bibliography{ref}
	}

\end{document}